\documentclass[a4paper,fleqn]{cas-sc}

\usepackage{amsmath,amssymb}
\usepackage{booktabs}
\usepackage{graphicx}
\usepackage{multirow}
\usepackage{array}
\usepackage{siunitx}
\usepackage{xurl}
\usepackage{capt-of}
\usepackage{placeins}
\usepackage{fancyvrb}
\fvset{fontsize=\footnotesize,frame=single,framesep=3pt,xleftmargin=0pt}
\usepackage[authoryear,longnamesfirst]{natbib}

\newcommand{\physopro}{Phys-OPRO}
\newcommand{\lesobj}{J_{\mathrm{LES}}}
\newcommand{\oscobj}{J_{\mathrm{osc}}}
\newcommand{\Ck}{C_k}
\newcommand{\Ce}{C_e}
\newcommand{\CLrms}{C_{L,\mathrm{rms}}}
\newcommand{\CD}{C_D}

\begin{document}
\let\WriteBookmarks\relax
\def\floatpagefraction{0.65}
\setcounter{topnumber}{3}
\setcounter{bottomnumber}{2}
\setcounter{totalnumber}{5}
\def\textpagefraction{0.15}
\def\topfraction{0.9}
\def\bottomfraction{0.8}

\shorttitle{Physics-labeled LLM prompting and numerical continuation for simulator calibration}
\shortauthors{Wang}

\title [mode = title]{When is a large language model a good calibration agent? Physics-labeled prompting for budgeted engineering simulator calibration}


 \author[1]{Ruo-Qian Wang}[orcid=0000-0003-2052-8948]
 \cormark[1]
 \ead{rq.wang@rutgers.edu}

 \affiliation[1]{organization={Department of Civil and Environmental Engineering, Rutgers University},
             addressline={500 Bartholomew Rd},
             city={Piscataway},
             state={NJ},
             postcode={08854},
             country={USA}}

 \cortext[cor1]{Corresponding author.}

\begin{abstract}
Calibrating an expensive engineering simulator means selecting uncertain model parameters so that simulated outputs match reference data, and the cost of doing so is dominated by the number of simulator runs the outer loop consumes. This paper asks whether a large language model (LLM), given a plain-text description of what is physically wrong with the current run, can select those runs more efficiently than a conventional derivative-free optimizer. The test cases are a large-eddy simulation of flow past a circular cylinder, calibrated against force-coefficient statistics, and an inexpensive damped-oscillator inverse problem. The physics-labeled prompt is compared with four ablated prompts, in which the diagnostic text is removed, anonymized, replaced by length-matched filler, or deliberately misdirected, and with Nelder--Mead, Gaussian-process Bayesian optimization, covariance-matrix adaptation and random search. All methods start from the same initial design. Because a language model returns a different answer each time it is called, the physics-labeled prompt is repeated 25 times and the ablated prompts 30 times, so that the comparison can be tested rather than merely described. The physical description produces a clear advantage over the ablated prompts during the first few simulator runs. Its first informed proposal improved on the best point of the shared starting design in 23 of 25 repeats, against 8 of 30 for the ablated prompts. In objective terms the median calibration error after five runs is 21.9 times lower than with the physics removed, 4.7 times lower after ten runs, and indistinguishable after fifteen. The repeats also quantify how variable the model itself is: identical prompts and identical starting points return five-seed medians spanning a factor of 9.8, which exceeds many of the effects this literature reports from five runs alone. Because the advantage does not persist, the LLM is used as a warm start and the accumulated runs are handed to a conventional optimizer at a fixed budget. The resulting workflow reaches the target accuracy in 9 simulator runs, against 12 for the LLM alone, 11--14 for the ablated prompts, 17 for Nelder--Mead and more than 60 for Bayesian optimization or random search. The eight runs saved relative to Nelder--Mead amount to roughly half an hour of solver time in the present case and to most of a working day on a simulator with hour-scale evaluations. A sensitivity analysis over all 3\,047 solver evaluations shows that the chosen force statistics constrain only one of the three calibrated parameters, which bounds how far any such ranking can be pushed. Physical descriptions buy an early sense of direction; they do not buy convergence.
\end{abstract}

\begin{highlights}
\item Physics descriptions in an LLM prompt cut early calibration error 22-fold
\item The advantage lasts about ten simulator runs, then disappears entirely
\item Repeats of one identical prompt give medians spanning a factor of 9.8
\item Handing LLM runs to a numerical optimizer hits the target in 9 runs against 17
\item Only one of three LES closure parameters is identifiable from force statistics
\end{highlights}


\maketitle

\section*{Nomenclature}
\noindent
\begin{tabular}{@{}ll@{}}
$x$ & bounded simulator parameter vector \\
$\mathcal{X}$ & feasible parameter set defined by the bounds \\
$J(x)$ & calibration objective (lower is better) \\
$\lesobj$, $\oscobj$ & LES and oscillator objectives, Eqs.~\eqref{eq:les-objective}, \eqref{eq:osc-objective} \\
$d(x)$, $q(x)$ & diagnostic vector and its text summary \\
$f_m$, $f_m^\star$ & simulated and target value of feature $m$ \\
$s_m$, $w_m$ & normalizing scale and weight of feature $m$ \\
$B$ & evaluation budget (number of simulator runs) \\
$B_{\mathrm{switch}}$ & budget at which the LLM hands over, $=15$ \\
$\tau$ & objective threshold used for budget-to-target \\
$n$ & number of repeated runs of one policy \\
$\nu$ & molecular kinematic viscosity, \si{\metre\squared\per\second} \\
$\Ck$, $\Ce$ & Smagorinsky closure coefficients \\
$\CD$, $\CLrms$ & mean drag and RMS lift coefficient \\
$k$, $c$ & oscillator stiffness and damping \\
CMA-ES & covariance-matrix adaptation evolution strategy \\
GP-BO & Gaussian-process Bayesian optimization \\
LES & large-eddy simulation \\
LLM & large language model \\
NM & Nelder--Mead \\
OPRO & optimization by prompting \\
\end{tabular}

\section{Introduction}

Engineering simulators encode decades of domain knowledge, but their practical value usually depends on calibration: selecting uncertain parameters so that simulated outputs match reference data. The cost of calibration is dominated by the outer loop that decides which simulation to run next. When one evaluation costs hours of solver time, the available budget may permit only a handful of decisions, and the quality of the first few determines whether a usable calibration is reached before the budget is exhausted. Calibration performance is therefore governed by more than final parameter accuracy. The number of evaluations spent sets the computational cost, and the quality of the early decisions sets how efficiently that cost is incurred.

\begin{figure}
    \centering
\includegraphics[width=\linewidth]{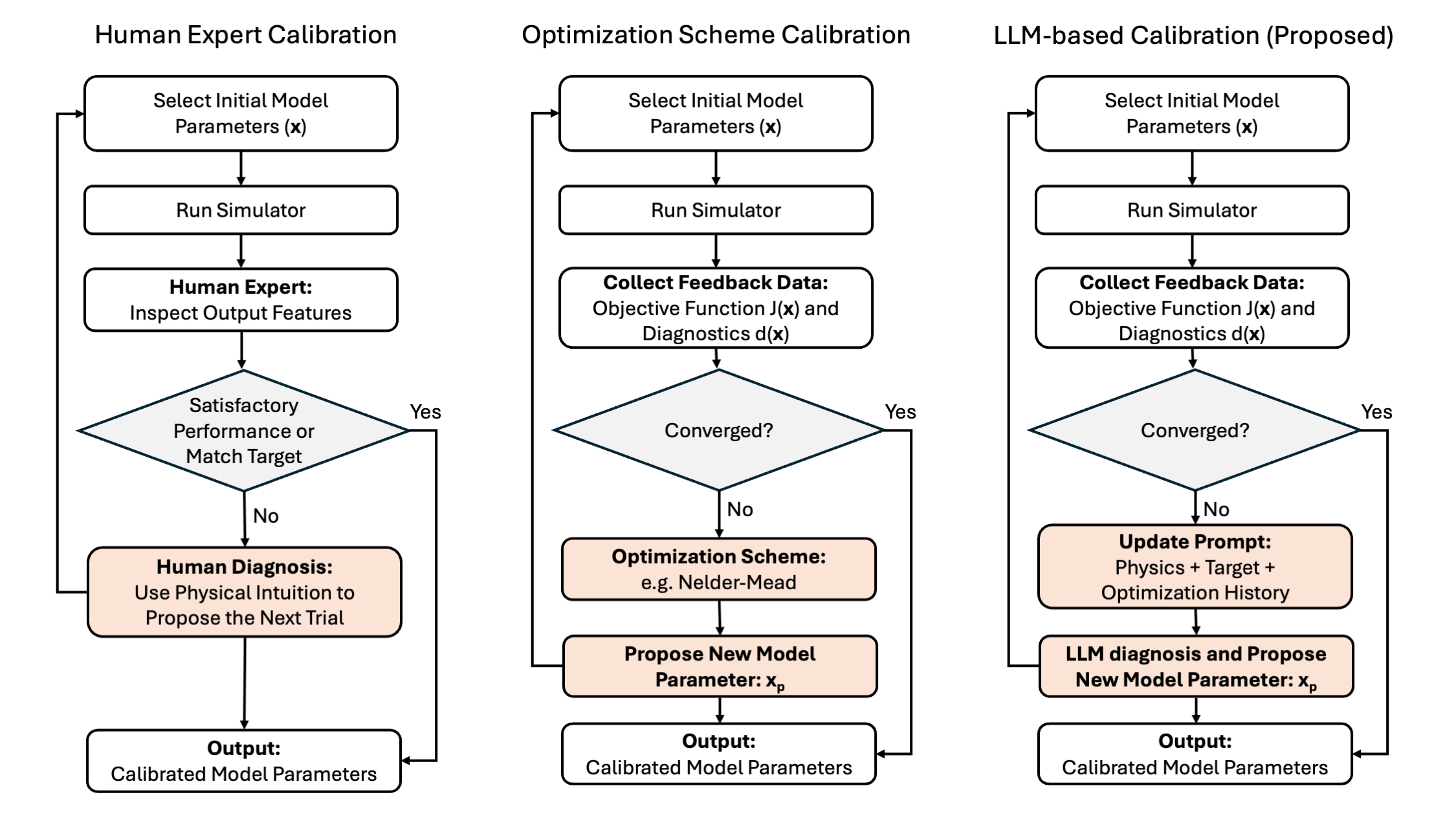}
    \caption{Three strategies for engineering simulator calibration under a limited evaluation budget: conventional numerical optimization, expert-guided calibration based on physical discrepancies, and a physics-labeled LLM proposal policy. The proposed workflow combines optimization history with physically interpretable diagnostics to select the next bounded simulator evaluation, and hands the accumulated evaluations to a numerical optimizer once the semantic channel is exhausted.}
    \label{fig:workflow}
\end{figure}

Two strategies have dominated practice, and they draw on different information (Fig.~\ref{fig:workflow}). The first is conventional derivative-free optimization. Nelder--Mead, covariance-matrix adaptation and Bayesian optimization \citep{Conn2009,NelderMead1965,Hansen2003,Snoek2012} are general and automatic, but they treat parameters as anonymous coordinates and infer search directions from the numerical history alone. The second is expert judgment. Human engineers inspect interpretable discrepancies, such as an overly damped oscillator, suppressed vortex shedding, an incorrect drag level or a shifted phase, and translate these into directed parameter changes. The expert loop uses information that the numerical methods discard, but it is labor-intensive, subjective and hard to reproduce, because the reasoning behind each adjustment is rarely written down \citep{Wang2011EngineeringModels}. Neither strategy serves the whole budget well. Model-based optimizers can learn useful relationships between parameters and objective values, but only once enough evaluations have accumulated to build a reliable approximation of the response, which leaves a gap in exactly the early regime where a directed guess would be most valuable. Expert judgment fills that gap, but it is itself uncertain and it scales poorly as the number of calibrated parameters grows. 

Large language models (LLMs) offer one route to closing that gap. An LLM conditions on natural language and can act as an iterative proposal engine, generating candidate solutions from a record of previous attempts \citep{Yang2024OPRO,Liu2024LLMOpt,Nie2024Directional,Zhang2024OPRO}. Two questions follow for engineering practice. The first is whether an LLM can serve as a calibration agent that consumes physically interpretable diagnostics and proposes the next simulator evaluation well enough to justify its place in the loop under a tight budget. The second is whether the physics competence that LLMs display when answering questions converts into useful action. Calibration is a stronger test than question answering, because the model must turn a description of a physical discrepancy into a specific intervention and is scored on whether the simulator output actually improves. Testing this requires a controlled comparison rather than a demonstration. The diagnostic language can be removed, anonymized, length-matched or deliberately misdirected while the numerical history is held fixed, which isolates the contribution of the physical content.

Because the use of LLMs for optimization and calibration is new to the engineering literature, the mechanism is stated here before the experiments. At each iteration the model receives a plain-text record of the parameter vectors already evaluated together with their objective values and, in the physics-labeled case, a few sentences naming which physical quantities are off and in which direction: that the wake is shedding too slowly, or that the lift amplitude is too weak. It returns one parameter vector in machine-readable form, which is clipped to the bounds and passed to the simulator (Fig.~\ref{fig:workflow}). No training, fitting or differentiation takes place. No surrogate response surface is built and no weights are updated, and the entire adaptation occurs inside a single prompt. The contrast with model-based optimization is instructive. Gaussian-process Bayesian optimization also proposes points from an evaluation history, but it does so by fitting an explicit probabilistic surrogate and maximizing an acquisition function over it, and that surrogate becomes informative only as evaluations accumulate. The LLM builds no such object. What it brings instead is whatever prior association it carries between a described physical symptom and a plausible parameter change. It also carries none of the guarantees of a numerical method: a language model can emit unparseable output, propose infeasible points or repeat a point it has already evaluated, and it has no convergence property.

This study therefore asks whether an LLM supplied with physical background information can be built into a more efficient calibration agent. The main contributions are as follows.
\begin{enumerate}
\item A bounded, physics-labeled LLM proposal policy is formulated for budgeted engineering simulator calibration and paired with a \emph{numerical continuation} step that transfers the accumulated evaluations to Nelder--Mead or Gaussian-process Bayesian optimization (GP-BO) at a fixed handoff budget.
\item The early advantage of the physical description is established by repeating each prompt many times rather than by comparing single trajectories. The clearest readout is the outcome of the first informed proposal, which improved on the shared starting design in 23 of 25 repeats of the physics-labeled prompt against 8 of 30 for the ablated prompts. In objective terms, the median best objective is $21.9$ times lower at five evaluations and $4.7$ times lower at ten evaluations, and the two are indistinguishable at fifteen.
\item The run-to-run variability of the LLM proposal policy is measured at fixed seed. Five arms running an identical policy on identical starting designs return five-seed medians spanning a factor of $9.8$ at five evaluations. This sets a resolution floor that any five-seed LLM-optimizer comparison must clear, and to our knowledge it has not previously been quantified on an engineering simulator.
\item The continuation step converts the transient advantage into a usable procedure. It matches the LLM early and Nelder--Mead late, and reaches an objective of $10^{-4}$ in a median of 9 evaluations against 12 for the pure LLM policy, 11--14 for the ablated prompts and 17 for Nelder--Mead.
\item A sensitivity analysis of the benchmark over all 3\,047 solver evaluations shows that only the molecular viscosity is identifiable from the four target statistics. This bounds which parts of the reported trajectories carry information and which sit inside the noise floor, and such a screen is recommended as routine practice before ranking optimizers on a calibration benchmark.
\end{enumerate}

\section{Related work}

\subsection{Engineering simulator calibration}

Calibration of computer models is a classical inverse problem in engineering, geoscience and uncertainty quantification \citep{KennedyOHagan2001,SungTuo2024}. When gradients are unavailable or unreliable, derivative-free optimization provides a general outer-loop strategy \citep{Conn2009}. Applications include climate-model parameter tuning \citep{Tett2017ClimateCalibration}, engineering analysis model validation \citep{Wang2011EngineeringModels} and ensemble-based calibration of complex numerical models \citep{Dunbar2022}. The advantage of these methods is their generality. Their limitation is that they usually operate on anonymous numerical coordinates and scalar objectives, discarding much of the interpretable physical information that engineers use when diagnosing a failed simulation.

This paper adopts two conventions from that literature. The first is that comparisons of derivative-free methods are conventionally reported through budget-to-target statistics and data profiles rather than through a single terminal objective \citep{more2009benchmarking}, because the quantity of practical interest is how many evaluations are needed to reach an acceptable accuracy. The second is that the identifiability of the calibrated parameters conditions any such comparison. If part of the parameter vector is weakly constrained by the chosen observables, the corresponding portion of the objective landscape is flat and optimizer rankings within it are not informative \citep{KennedyOHagan2001}.

\subsection{LLMs as proposal mechanisms}

Optimization by prompting treats an LLM as an iterative proposal engine that receives a history of attempted solutions and returns a new candidate \citep{Yang2024OPRO}. Related work uses LLMs to guide evolutionary search or to exploit directional feedback \citep{Liu2024LLMOpt,Nie2024Directional}. Recent studies also point to limitations: performance may depend on model size, prompt template, task distribution, and whether the feedback contains actionable directions \citep{Zhang2024OPRO}. For engineering applications this motivates careful benchmarking against derivative-free methods and prompt ablations, rather than broad claims that an LLM is itself a superior optimizer.

One issue is largely absent from this literature, namely the variability of the proposals themselves. Trajectories on an expensive simulator are typically reported over a handful of seeds, and the seed is usually assumed to control the whole experiment, which it does not. A stochastic decoder returns different proposals on repeated calls with an identical prompt, so the seed fixes the initial design and the numerical baselines but leaves the language model free to vary. An LLM policy is therefore a distribution over trajectories even when the environment is fully seeded, and a curve obtained from a single arm confounds policy quality with decoder sampling. The number of repeats required to resolve a given effect follows from the width of that distribution. This width is measured directly here and used to judge how much of the observed benchmark separation can be trusted.

\subsection{LES calibration as an engineering test case}

The main engineering case in this paper is calibration of a Smagorinsky-type LES closure \citep{Smagorinsky1963,Lilly1967} for flow past a circular cylinder, a canonical vortex-shedding problem \citep{Williamson1996}. The simulator is run through OpenFOAM \citep{Weller1998}; the outer loop varies closure and viscosity parameters and compares force-coefficient statistics against reference data. This setting suits the present question because the simulator output contains meaningful diagnostics while each evaluation is costly enough that early-budget behavior matters. As Section~\ref{sec:identifiability} shows, it is also a setting in which the observables constrain the closure coefficients only weakly. That behavior is typical of force-statistics-based calibration, and it is the reason the sensitivity analysis is reported alongside the optimizer comparison.

\section{Physics-labeled LLM calibration with numerical continuation}

\subsection{Calibration problem}

Let $x\in\mathcal{X}$ denote bounded simulator parameters, $S(x)$ the simulator output, and $y^\star$ the target observation or reference output. The calibration objective is
\begin{equation}
    \min_{x\in\mathcal{X}} J(x),
\end{equation}
where $J$ is available only by running the simulator and post-processing its output. In addition to $J(x)$, the evaluator may return a diagnostic vector $d(x)$ and a short text summary $q(x)$ describing the physical discrepancy. The prompt-based proposal policy receives a history
\begin{equation}
    \mathcal{H}_t = \{x_i, J(x_i), d(x_i), q(x_i)\}_{i=1}^{t}
\end{equation}
and returns a new bounded candidate $x_{t+1}$.

The physics-labeled LLM policy used here is deliberately simple. The prompt lists the parameter bounds, the ten best previous trials sorted by objective, and a concise physical diagnostic block for the incumbent. The model is required to return a machine-readable parameter vector, and invalid or out-of-bounds proposals are clipped or retried by the driver script. The full physics-labeled policy is referred to as \physopro{} where a short name is convenient, although the paper emphasizes the workflow rather than the method name. The complete prompt text is reproduced in Appendix~\ref{app:prompts}, and the parsing and retry logic are given in Appendix~\ref{app:harness}.

\subsection{Numerical continuation}
\label{sec:continuation}

An advantage confined to the first few evaluations is of limited use unless the workflow specifies when to stop relying on it. If physical semantics help only while the numerical history is uninformative, the deliverable is a warm-start rule rather than a stand-alone optimizer. The policy described above is treated accordingly. Once the numerical history is informative, an LLM has no mechanism for local refinement and, as Section~\ref{sec:hygiene} shows, spends an increasing share of its budget re-proposing points it has already evaluated.

A two-stage policy is therefore defined. For the first $B_{\mathrm{switch}}$ evaluations the physics-labeled policy proposes points as above. At $B_{\mathrm{switch}}$ the entire evaluated set $\{(x_i,J(x_i))\}_{i=1}^{B_{\mathrm{switch}}}$ is handed to a conventional derivative-free optimizer, which continues to the full budget. Two continuations are considered: Nelder--Mead restarted from the incumbent best point, and GP-BO initialized with the complete evaluated set as its design. The value $B_{\mathrm{switch}}=15$ is used throughout, chosen \emph{a priori} as the harness default and not tuned on the results.

This construction has a useful experimental consequence. For $B\le B_{\mathrm{switch}}$ the hybrid arms and the standalone physics-labeled arm execute an identical policy on an identical initial design. Any difference between them at those budgets is therefore attributable to proposal variability alone, and pooling them yields repeated runs of a single policy. This property is used in Sections~\ref{sec:prompt-controls} and~\ref{sec:variance}.

\subsection{Prompt controls}

To separate physical semantics from generic prompt effects, several controls are evaluated (Table~\ref{tab:prompt-variants}). The no-diagnostic prompt removes the physical diagnostic text and leaves only the parameter and objective history. The anonymized prompt retains the numeric metric values but replaces parameter and metric names by anonymous symbols. The length-matched prompt replaces the diagnostic language with inert filler text of comparable length while retaining the numeric values. The permuted-semantics prompt intentionally mismatches or misdirects the physical labels. These controls are imperfect but useful: if the full prompt wins only because it is longer or because it exposes more numbers, then the length-matched and anonymized prompts should perform similarly to it.

\begin{table}[t]
\centering
\caption{Prompt variants used to separate physical semantics from generic prompt structure. Exact template text is given in Appendix~\ref{app:prompts}.}
\label{tab:prompt-variants}
\small
\begin{tabular}{p{0.28\linewidth}p{0.62\linewidth}}
\toprule
Variant & Information given to the proposal model \\
\midrule
physics-labeled LLM & Parameter names, units, bounded ranges, objective history, and physical discrepancy diagnostics. \\
no diagnostics & Parameter and objective history only; diagnostic text is removed. \\
anonymized metrics & Numeric diagnostic values are retained but labels are replaced by anonymous metric names and parameters by $p_1,\dots,p_d$. \\
length-matched text & A text block of comparable length with inert filler tokens and numeric values, without valid physical labels or units. \\
permuted semantics & Physical labels are intentionally mismatched or directionally misleading. \\
\bottomrule
\end{tabular}
\end{table}

\subsection{Comparison questions}

The evaluation keeps three comparisons separate. The \emph{prompt-ablation} comparison holds the LLM proposal mechanism fixed while changing the semantic content of the prompt, and asks whether physically meaningful information contributes to performance. The \emph{optimizer} comparison contrasts the LLM-based policies with standalone random search, Nelder--Mead, GP-BO and covariance-matrix adaptation where applicable, and asks whether the policies are sample-efficient as optimizers. The \emph{continuation} comparison contrasts the two-stage policy against both of its components. Keeping the three separate prevents performance supplied by a different optimizer from being mistaken for evidence about the value of physical prompting.

\section{Experimental setup}

\subsection{LES cylinder-wake calibration}

\begin{figure}
    \centering
    \includegraphics[width=0.98\linewidth]{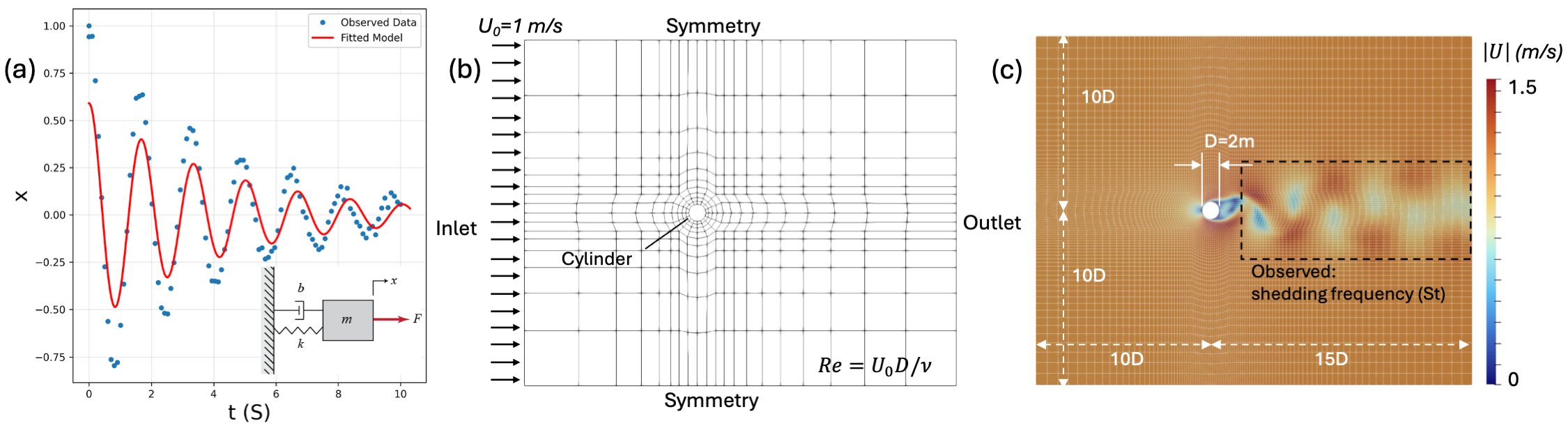}
    \caption{Engineering benchmark overview. The study evaluates budgeted simulator calibration on two settings: (a) a damped-oscillator inverse problem and (b, c) calibration of a two-dimensional cylinder-wake LES against a higher-resolution reference of the same solver. The LES task calibrates molecular viscosity $\nu$ and Smagorinsky closure parameters $\Ck$ and $\Ce$ using wake statistics including shedding frequency, mean drag, RMS lift, and phase.}
    \label{fig:benchmark_overview}
\end{figure}

The primary engineering benchmark calibrates a Smagorinsky LES closure for two-dimensional incompressible flow past a circular cylinder (Figs.~\ref{fig:benchmark_overview}b and \ref{fig:benchmark_overview}c). The OpenFOAM case uses \texttt{pisoFoam}. The calibrated parameters are the molecular viscosity $\nu$, the Smagorinsky coefficient $\Ck$ and the dissipation coefficient $\Ce$, with bounded ranges
\begin{equation}
    \nu\in[0.003,0.015],\quad \Ck\in[0.005,0.25],\quad \Ce\in[0.20,2.50].
    \label{eq:bounds}
\end{equation}
The reference target is extracted from the force coefficients of a higher-resolution reference run of the same solver, so the benchmark is a self-consistent identifiability problem and not a validation of two-dimensional LES against physical turbulence; this framing is revisited in Section~\ref{sec:limitations}. The evaluator computes four statistics over the window $t\in[250,350]$: the dominant frequency of the drag signal, which equals twice the lift-shedding frequency and is used here as a shedding-frequency proxy; the RMS lift coefficient $\CLrms$; the mean drag coefficient $\CD$; and the drag phase at $t=250$. The normalized objective is
\begin{equation}
    \lesobj(x)=\frac{\sum_{m} w_m \left(\frac{f_m(x)-f_m^\star}{s_m}\right)^2}{\sum_m w_m},
    \label{eq:les-objective}
\end{equation}
where phase errors use wrapped angular differences. All four weights $w_m$ are set to one, and the scales $s_m$ are the target values themselves except for the phase, which is scaled by $\pi$; numerical values are given in Appendix~\ref{app:features}. The same scalar objective is used for every method. The distinguishing feature of the physics-labeled LLM is that it additionally receives the vector of component discrepancies and their physical interpretations in the prompt. Failed solver runs receive a penalty of $10^{9}$.

Two LES batches were run. A 60-evaluation batch covers the physics-labeled policy, the two continuations, the no-diagnostic and permuted-semantics controls, random search, Nelder--Mead and GP-BO. A 30-evaluation batch covers the physics-labeled policy, the Nelder--Mead continuation and all four semantic controls. Both batches use the same five seeds, $\{3554190, 4866770, 5280212, 8667532, 9487190\}$, and within each seed every arm begins from a byte-identical four-point initial design (Appendix~\ref{app:harness}). Together the two batches comprise 3\,300 logged evaluations, of which 3\,047 required a solver run, totalling \SI{226}{\hour} of solver time at a mean of \SI{267}{\second} per evaluation.

\subsection{Randomized damped oscillator}

The secondary benchmark is a damped oscillator inverse problem (Fig.~\ref{fig:benchmark_overview}a). The simulator solves
\begin{equation}
    \ddot{u}(t)+c\dot{u}(t)+ku(t)=0,
\end{equation}
with fixed initial conditions and unit mass. For each seed the hidden target parameters $(k^\star,c^\star)$ are sampled inside the calibration bounds, observations are generated with Gaussian noise, and each method minimizes the mean-squared displacement error
\begin{equation}
    \oscobj(k,c)=\frac{1}{N}\sum_{j=1}^{N}\left(u(t_j;k,c)-u_j^{\mathrm{obs}}\right)^2 .
    \label{eq:osc-objective}
\end{equation}
The oscillator is inexpensive, so it is run over 30 randomized trajectories per method at budgets of 15, 30 and 60 evaluations. Because the targets are resampled per seed, this benchmark also removes the possibility that the LLM recalls a canonical parameter value, which is a live concern for the LES case (Section~\ref{sec:discussion}).

\subsection{Performance measures}
\label{sec:protocol}

Every method is summarized by the best objective found so far at each prefix budget $B$, which is the calibration error an engineer would have in hand after $B$ simulator runs. Three quantities are reported.

\emph{Best objective against budget.} The median over repeated runs of the best-so-far objective at $B=5$, 10, 15, 30 and 60.

\emph{Evaluations to target.} Following standard practice in derivative-free optimization \citep{more2009benchmarking}, the number of simulator runs needed to first reach a fixed objective threshold $\tau$, reported as a median together with data profiles giving the fraction of seeds that reach $\tau$ by each budget. The thresholds $\tau\in\{10^{-3},2\times10^{-4},10^{-4}\}$ span the range between an obviously inadequate calibration and the empirical floor of the benchmark. This is the decision-relevant number for practice, because it converts directly into solver hours.

\emph{First-proposal hit rate.} The fraction of runs in which the first method-specific proposal, evaluation 5, produced a lower objective than the best of the four initialization points that every method was given. This is a plain count of successes and failures on the single decision that the diagnostic text is meant to influence, and it does not depend on the size of any individual objective value.

A single terminal objective is deliberately avoided as the headline number. Section~\ref{sec:identifiability} shows that beyond roughly fifteen evaluations all methods have entered a flat region of the objective in which the differences between them are smaller than the benchmark can resolve.

\subsection{Repeated runs}
\label{sec:repeats}

Because the language model returns a different answer each time it is called, a single trajectory carries little information. Section~\ref{sec:continuation} noted that the standalone physics-labeled arm and both continuation arms execute exactly the same policy for $B\le15$. Pooling the three such arms of the 60-evaluation batch with the two of the 30-evaluation batch therefore gives $n=25$ repeated runs of one policy at each of $B=5$, 10 and 15. The same pooling gives $n=10$ repeats each for the no-diagnostic and permuted-semantics controls and $n=5$ each for the anonymized and length-matched controls, 30 in total. Repetition is what converts the central question from a description of five trajectories into a testable comparison, and it is the reason the prompt contrast can be reported with an uncertainty range rather than as a ranking of medians.

\subsection{Assessing the differences}
\label{sec:inference}

The comparisons in Section~\ref{sec:results} use three standard checks, chosen so that no assumption about the shape of the error distribution is required. LES objective values are strongly skewed, since an occasional poor run produces a value orders of magnitude above the rest, so medians are reported instead of means and differences between arms are expressed as ratios of medians.

The uncertainty on each median is obtained by resampling the repeated runs with replacement many times and reading off the central 95\% of the resulting medians. This is the standard bootstrap interval, and it is written as a 95\% confidence interval in the tables.

Whether two arms differ by more than chance is assessed by repeatedly reshuffling the arm labels among the pooled runs and counting how often a reshuffle produces a ratio at least as extreme as the observed one. This is a permutation test, and the reported $p$ is that fraction. A value of $p<10^{-4}$ therefore means that fewer than one reshuffle in ten thousand reproduced the observed gap by chance; a value near $0.1$ means that roughly one reshuffle in ten did, so the observed gap would be unremarkable.

The first-proposal hit rates are simple counts of successes out of trials and are compared with Fisher's exact test, which is the standard test for a two-by-two table of counts. On the oscillator, where every method is run on the same 30 randomized targets, methods are compared trajectory by trajectory using the paired Wilcoxon signed-rank test, which ranks the per-trajectory differences instead of averaging them.

Resampling counts, the treatment of multiple comparisons and the software used are collected in Appendix~\ref{app:stats}, so that the results sections can be read without them.

\section{Results}
\label{sec:results}

\subsection{Effect of the physical diagnostics}
\label{sec:prompt-controls}

\begin{figure}
    \centering
\includegraphics[width=\linewidth]{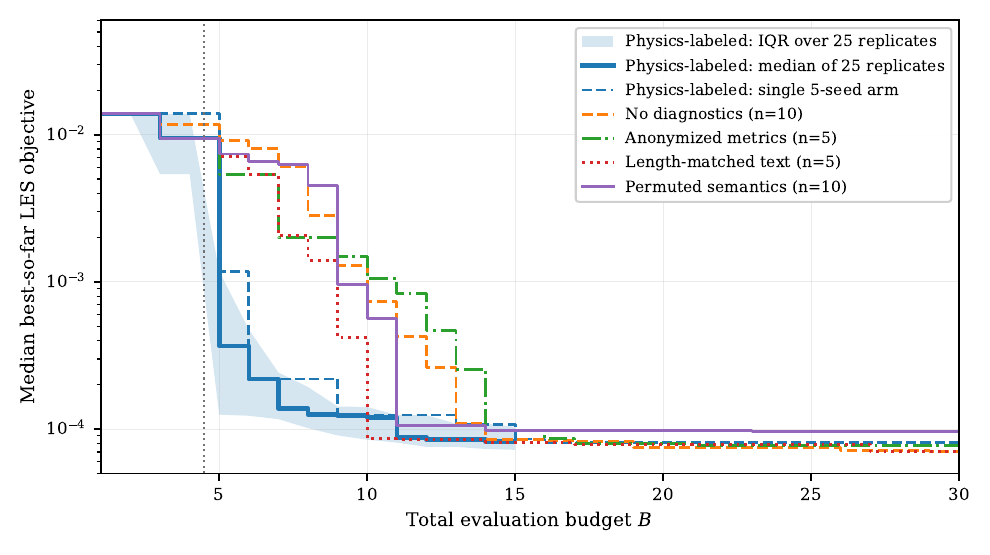}
    \caption{LES prompt-control batch. The heavy line and shaded band give the median and interquartile range of the physics-labeled policy over 25 repeated runs at matched seeds. The dashed blue line is a single five-seed arm of the same policy, included to show how far one such arm can sit from the median of the repeats. Control curves are medians over the runs listed in Table~\ref{tab:les-controls}. The dotted vertical line separates the shared four-point initialization from the first method-specific proposal.}
    \label{fig:les-controls}
\end{figure}

Two views of the same comparison are presented. The first is a count of how often the first informed proposal helped at all (Table~\ref{tab:first-proposal}). The second is the size of the resulting difference in calibration error (Table~\ref{tab:les-controls} and Fig.~\ref{fig:les-controls}).

\subsubsection{The first informed proposal}

Every arm was handed the same four initialization points, so evaluation 5 is the first decision that the diagnostic text can influence. The physics-labeled prompt produced a better point than the best of those four in 23 of its 25 repeated runs, a hit rate of 92\%. The ablated prompts did so in 8 of 30 runs, a hit rate of 27\%, and the difference between the two rates is far outside what chance would produce (Fisher exact $p=1.3\times10^{-6}$). Twenty-two of the thirty ablated runs therefore did not merely make a poor proposal; they failed to improve on a design point they had been given for free. Because this readout is a count, it is insensitive to the size of any single objective value, and it isolates the one decision the diagnostic text is supposed to influence. The numerical baselines appear in the same table for reference, although Nelder--Mead is structurally handicapped at this budget: with three parameters it spends evaluation 5 completing its initial simplex.

\subsubsection{Size and persistence of the effect}

At five evaluations the physics-labeled prompt reaches a median best objective of $3.68\times10^{-4}$ against $8.03\times10^{-3}$ for the pooled controls, a factor of $21.9$ with $p<10^{-4}$. Each control separates individually: no diagnostics by a factor of $25.0$ ($p=4.5\times10^{-4}$), permuted semantics by $20.1$ ($p=4.0\times10^{-4}$), length-matched text by $19.5$ ($p=0.020$) and anonymized metrics by $14.7$ ($p=0.036$). Four comparisons are made at once here, which raises the chance of one appearing significant by accident; correcting for this (Appendix~\ref{app:stats}) leaves every conclusion in this section unchanged.

The length-matched control is the informative one. It carries the same number of characters and the same numbers as the physics-labeled block and differs only in that its words carry no physical meaning (Appendix~\ref{app:prompts}), so its separation by a factor of $19.5$ cannot be an effect of prompt length. The anonymized and no-diagnostic blocks are substantially shorter than the physics-labeled block, so those two comparisons remove text as well as labels and are correspondingly less clean.

At ten evaluations the advantage over the pooled controls has fallen to a factor of $4.7$ ($p<10^{-4}$) and is no longer uniform. The length-matched control has caught up and is nominally ahead ($8.62\times10^{-5}$ against $1.20\times10^{-4}$, $p=0.71$), while the no-diagnostic, anonymized and permuted controls remain behind by factors of $6.1$, $8.8$ and $4.7$.

At fifteen evaluations the advantage has disappeared. The pooled ratio is $1.12$ with $p=0.099$, no individual control is separated at any conventional level, and the interval for the physics-labeled policy, $[7.27,9.00]\times10^{-5}$, overlaps that of every control.

Both readouts give the same picture: a large benefit at the first informed proposal, decaying with each subsequent evaluation, and absent by the fifteenth. The permuted-semantics control, which supplies deliberately wrong physics, converges to the same region as the rest. What the physical labels appear to supply is therefore a head start rather than sustained physical reasoning.

\begin{table*}[t]
\centering
\caption{LES prompt-control comparison on pooled repeated runs. Lower is better. The physics-labeled policy is repeated $n=25$ times at matched seeds (Section~\ref{sec:repeats}); controls are repeated as indicated. Intervals are $95\%$ bootstrap confidence intervals on the median. ``Ratio'' is the control median divided by the physics-labeled median, so values above one favor the physics-labeled prompt. $p$ is the permutation test described in Section~\ref{sec:inference}. The wide control intervals at $B=10$ and $B=15$ reflect the small number of repeats available for the anonymized and length-matched arms and are a direct expression of the variability quantified in Table~\ref{tab:variance}.}
\label{tab:les-controls}
\small
\begin{tabular}{llrrrr}
\toprule
$B$ & Arm & $n$ & Median $\lesobj$ & 95\% CI & Ratio ($p$) \\
\midrule
\multirow{6}{*}{5}
& Physics-labeled LLM & 25 & $3.68\times10^{-4}$ & $[1.4\times10^{-4},\ 9.8\times10^{-4}]$ & --- \\
& No diagnostics & 10 & $9.19\times10^{-3}$ & $[5.4\times10^{-3},\ 1.4\times10^{-2}]$ & $25.0$ $(4.5\times10^{-4})$ \\
& Anonymized metrics & 5 & $5.41\times10^{-3}$ & $[1.0\times10^{-4},\ 1.4\times10^{-2}]$ & $14.7$ $(0.036)$ \\
& Length-matched text & 5 & $7.16\times10^{-3}$ & $[8.0\times10^{-4},\ 1.4\times10^{-2}]$ & $19.5$ $(0.020)$ \\
& Permuted semantics & 10 & $7.40\times10^{-3}$ & $[3.1\times10^{-3},\ 1.4\times10^{-2}]$ & $20.1$ $(4.0\times10^{-4})$ \\
& \textit{All controls pooled} & 30 & $8.03\times10^{-3}$ & $[5.4\times10^{-3},\ 9.5\times10^{-3}]$ & $\mathbf{21.9}$ $(<10^{-4})$ \\
\midrule
\multirow{6}{*}{10}
& Physics-labeled LLM & 25 & $1.20\times10^{-4}$ & $[8.5\times10^{-5},\ 1.3\times10^{-4}]$ & --- \\
& No diagnostics & 10 & $7.33\times10^{-4}$ & $[1.0\times10^{-4},\ 8.9\times10^{-3}]$ & $6.09$ $(2.0\times10^{-4})$ \\
& Anonymized metrics & 5 & $1.06\times10^{-3}$ & $[9.0\times10^{-5},\ 1.2\times10^{-2}]$ & $8.83$ $(0.0024)$ \\
& Length-matched text & 5 & $8.62\times10^{-5}$ & $[6.9\times10^{-5},\ 1.2\times10^{-2}]$ & $0.72$ $(0.71)$ \\
& Permuted semantics & 10 & $5.63\times10^{-4}$ & $[1.0\times10^{-4},\ 5.1\times10^{-3}]$ & $4.68$ $(1.5\times10^{-4})$ \\
& \textit{All controls pooled} & 30 & $5.71\times10^{-4}$ & $[1.1\times10^{-4},\ 5.2\times10^{-3}]$ & $\mathbf{4.74}$ $(<10^{-4})$ \\
\midrule
\multirow{6}{*}{15}
& Physics-labeled LLM & 25 & $8.22\times10^{-5}$ & $[7.3\times10^{-5},\ 9.0\times10^{-5}]$ & --- \\
& No diagnostics & 10 & $8.52\times10^{-5}$ & $[6.7\times10^{-5},\ 4.3\times10^{-3}]$ & $1.04$ $(0.26)$ \\
& Anonymized metrics & 5 & $9.85\times10^{-5}$ & $[6.9\times10^{-5},\ 9.2\times10^{-3}]$ & $1.20$ $(0.19)$ \\
& Length-matched text & 5 & $8.11\times10^{-5}$ & $[6.6\times10^{-5},\ 1.0\times10^{-2}]$ & $0.99$ $(0.67)$ \\
& Permuted semantics & 10 & $9.76\times10^{-5}$ & $[7.1\times10^{-5},\ 2.9\times10^{-3}]$ & $1.19$ $(0.10)$ \\
& \textit{All controls pooled} & 30 & $9.17\times10^{-5}$ & $[7.1\times10^{-5},\ 4.0\times10^{-4}]$ & $\mathbf{1.12}$ $(0.099)$ \\
\bottomrule
\end{tabular}
\end{table*}

\begin{table}[t]
\centering
\caption{Outcome of the \emph{first} method-specific proposal: the fraction of runs in which evaluation 5 improved on the best of the shared four-point initialization. This is a count of successes on the single decision the diagnostic text is meant to influence, so unlike the objective ratios it does not depend on the size of individual $\lesobj$ values. Numerical baselines are shown for reference but are structurally handicapped here: with three parameters, Nelder--Mead spends evaluation 5 completing its simplex.}
\label{tab:first-proposal}
\small
\begin{tabular}{lrr}
\toprule
Arm & Improved / runs & Rate \\
\midrule
Physics-labeled LLM & 23/25 & $\mathbf{92\%}$ \\
\midrule
No diagnostics & 3/10 & $30\%$ \\
Anonymized metrics & 2/5 & $40\%$ \\
Length-matched text & 1/5 & $20\%$ \\
Permuted semantics & 2/10 & $20\%$ \\
\textit{All semantic controls} & 8/30 & $\mathit{27\%}$ \\
\midrule
Nelder--Mead & 0/5 & $0\%$ \\
GP-BO & 2/5 & $40\%$ \\
Random search & 3/5 & $60\%$ \\
\bottomrule
\end{tabular}
\end{table}

\subsection{Run-to-run variability of the proposal policy}
\label{sec:variance}

The same repeat structure that supports Table~\ref{tab:les-controls} also measures how reproducible a single five-seed LLM arm is. Table~\ref{tab:variance} lists the five-seed median of each of the five arms that run the identical physics-labeled policy on the identical seeds. Any difference between these rows is caused by the language model alone, since the policy, the seeds and the initial designs are the same.

\begin{table}[t]
\centering
\caption{Run-to-run variability of the physics-labeled policy. All five arms execute an identical policy on an identical initial design at the same five seeds for $B\le15$, so the differences between rows are caused entirely by the language model returning different answers to the same prompt. Each row is the median over five seeds, that is, exactly the statistic a conventional five-seed study would report.}
\label{tab:variance}
\small
\begin{tabular}{lrrr}
\toprule
Repeat arm & $B=5$ & $B=10$ & $B=15$ \\
\midrule
60-eval batch, no handoff        & $1.21\times10^{-3}$ & $1.41\times10^{-4}$ & $8.23\times10^{-5}$ \\
60-eval batch, NM continuation   & $3.56\times10^{-4}$ & $1.24\times10^{-4}$ & $8.36\times10^{-5}$ \\
60-eval batch, GP-BO continuation& $1.24\times10^{-4}$ & $8.41\times10^{-5}$ & $7.32\times10^{-5}$ \\
30-eval batch, no handoff        & $1.18\times10^{-3}$ & $1.25\times10^{-4}$ & $8.57\times10^{-5}$ \\
30-eval batch, NM continuation   & $3.68\times10^{-4}$ & $9.11\times10^{-5}$ & $8.14\times10^{-5}$ \\
\midrule
Spread (max/min) & $\mathbf{9.8\times}$ & $1.7\times$ & $1.2\times$ \\
Pooled median ($n=25$) & $3.68\times10^{-4}$ & $1.20\times10^{-4}$ & $8.22\times10^{-5}$ \\
Pooled IQR & $[1.3\times10^{-4},\,1.2\times10^{-3}]$ & $[0.9,1.4]\times10^{-4}$ & $[0.7,1.0]\times10^{-4}$ \\
\bottomrule
\end{tabular}
\end{table}

At five evaluations the five medians span a factor of $9.8$, from $1.24\times10^{-4}$ to $1.21\times10^{-3}$. Within a single seed the spread is wider still, reaching a factor of 92 on seed 4866770. The variability falls with budget, to $1.7\times$ at ten evaluations and $1.2\times$ at fifteen, because the best-so-far objective is a running minimum and its scatter is progressively absorbed as evaluations accumulate.

This result has two practical implications. First, a five-seed comparison of LLM policies at low budget on this benchmark cannot resolve anything smaller than about an order of magnitude, so a reported difference below that cannot safely be attributed to the manipulation under study. The differences reported in Table~\ref{tab:les-controls} at $B=5$ and $B=10$ clear this floor because they are computed on pooled repeats. The near-unity ratios at $B=15$ do not clear it, and they are accordingly reported as an absence of evidence rather than as evidence of equivalence. Second, repeating the policy is not interchangeable with adding seeds. Seeds control the initial design and the numerical baselines but not the language model, so raising the seed count without repeating the policy reduces one source of scatter and leaves the other untouched. Both counts should therefore be reported explicitly in LLM-optimizer studies.

\subsection{Optimizer comparison and numerical continuation}
\label{sec:optimizer}

\begin{figure}
    \centering
\includegraphics[width=\linewidth]{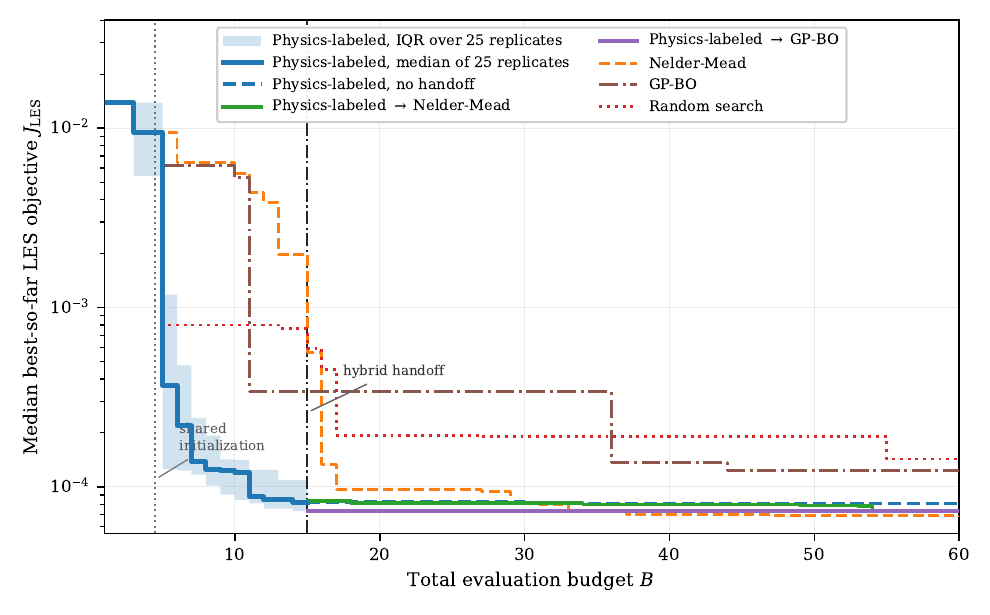}
    \caption{LES comparison with standalone optimizers and numerical continuation. The heavy line and band give the median and interquartile range of the physics-labeled policy over 25 repeated runs up to the handoff budget; after the handoff the three continuations separate. Step curves are medians over five shared seeds. The dotted line marks the end of the shared four-point initialization and the dash-dotted line the handoff budget $B_{\mathrm{switch}}=15$.}
    \label{fig:les-budget}
\end{figure}

Table~\ref{tab:les-main} and Fig.~\ref{fig:les-budget} compare the LLM-based policies with the standalone numerical optimizers on the five shared seeds of the 60-evaluation batch.

The standalone physics-labeled policy is far ahead of Nelder--Mead and GP-BO through the first fifteen evaluations. At $B=10$ its median is $1.41\times10^{-4}$ against $5.59\times10^{-3}$ and $5.29\times10^{-3}$ respectively. It then stops improving, ending at $8.10\times10^{-5}$, whereas Nelder--Mead reaches $6.88\times10^{-5}$ by $B=60$. This crossover is the pattern that motivates the continuation step.

Both continuations behave as intended. Through $B=15$ they are the physics-labeled policy and inherit its early behavior. After the handoff they resume descending: the Nelder--Mead continuation improves from $8.36\times10^{-5}$ at $B=15$ to $7.32\times10^{-5}$ at $B=60$, beating the pure LLM policy on all five seeds, and the GP-BO continuation reaches $7.32\times10^{-5}$ against $1.24\times10^{-4}$ for standalone GP-BO. Neither continuation is worse than its own numerical component at any reported budget. In engineering terms, the handoff costs nothing and removes the stall.

\begin{table*}[t]
\centering
\caption{Median best-so-far LES objective over five shared seeds, 60-evaluation batch. Lower is better. Arms above the rule use the physics-labeled prompt and are identical for $B\le15$ by construction; entries at $B=5$ and $B=10$ in those rows are therefore repeats of one policy rather than distinct methods, and Table~\ref{tab:variance} gives their spread. Section~\ref{sec:identifiability} shows that differences at $B=30$ and $B=60$ lie within the resolution of the benchmark and should not be read as a ranking.}
\label{tab:les-main}
\small
\begin{tabular}{lrrrrr}
\toprule
Method & $B=5$ & $B=10$ & $B=15$ & $B=30$ & $B=60$ \\
\midrule
Physics-labeled LLM, no handoff & $1.21\times10^{-3}$ & $1.41\times10^{-4}$ & $8.23\times10^{-5}$ & $8.10\times10^{-5}$ & $8.10\times10^{-5}$ \\
Physics-labeled $\rightarrow$ Nelder--Mead & $3.56\times10^{-4}$ & $1.24\times10^{-4}$ & $8.36\times10^{-5}$ & $8.11\times10^{-5}$ & $7.32\times10^{-5}$ \\
Physics-labeled $\rightarrow$ GP-BO & $1.24\times10^{-4}$ & $8.41\times10^{-5}$ & $7.32\times10^{-5}$ & $7.32\times10^{-5}$ & $7.32\times10^{-5}$ \\
\midrule
No diagnostics & $8.90\times10^{-3}$ & $9.01\times10^{-4}$ & $8.61\times10^{-5}$ & $7.19\times10^{-5}$ & $6.87\times10^{-5}$ \\
Permuted semantics & $5.41\times10^{-3}$ & $2.13\times10^{-4}$ & $7.06\times10^{-5}$ & $6.81\times10^{-5}$ & $6.81\times10^{-5}$ \\
Nelder--Mead & $9.48\times10^{-3}$ & $5.59\times10^{-3}$ & $5.64\times10^{-4}$ & $8.23\times10^{-5}$ & $6.88\times10^{-5}$ \\
GP-BO & $6.16\times10^{-3}$ & $5.29\times10^{-3}$ & $3.38\times10^{-4}$ & $3.38\times10^{-4}$ & $1.24\times10^{-4}$ \\
Random search & $7.99\times10^{-4}$ & $7.99\times10^{-4}$ & $5.89\times10^{-4}$ & $1.91\times10^{-4}$ & $1.43\times10^{-4}$ \\
\bottomrule
\end{tabular}
\end{table*}

The table needs two qualifications. The first concerns random search, which is competitive with the physics-labeled policy at $B=5$ ($7.99\times10^{-4}$ against a pooled $3.68\times10^{-4}$); only $56\%$ of the physics-labeled repeats beat the random-search median at that budget. The advantage established in Section~\ref{sec:prompt-controls} is therefore an advantage over other \emph{prompts}, and not an advantage over undirected sampling in a three-dimensional box at a single proposal. The second concerns the large-budget columns, where the no-diagnostic and permuted-semantics controls end below the physics-labeled policy at $B=60$. This is not evidence that removing physics helps. As Section~\ref{sec:identifiability} shows, all four values lie within a $20\%$ band, in a region where a $1.2\%$ change in objective spans the whole feasible range of two of the three parameters.

\subsection{Budget to target}
\label{sec:budget-to-target}

\begin{figure}
    \centering
\includegraphics[width=\linewidth]{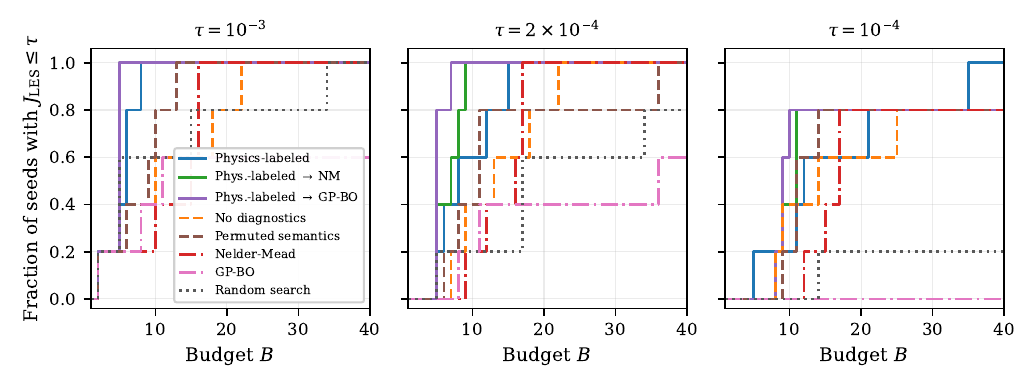}
    \caption{Data profiles for the LES benchmark: fraction of the five seeds that have reached objective $\tau$ by budget $B$, for three thresholds. The physics-labeled policies and their continuations dominate at every threshold; standalone GP-BO and random search fail to reach the tightest threshold on most seeds within the budget.}
    \label{fig:les-profile}
\end{figure}

Table~\ref{tab:budget} and Fig.~\ref{fig:les-profile} report the quantity that matters most in practice: how many simulator evaluations are needed to reach a given objective.

\begin{table}[t]
\centering
\caption{Median number of LES evaluations required to first reach objective $\tau$, over five shared seeds. Entries of $>60$ indicate that the median seed did not reach $\tau$ within the 60-evaluation budget.}
\label{tab:budget}
\small
\begin{tabular}{lrrr}
\toprule
Method & $\tau=10^{-3}$ & $\tau=2\times10^{-4}$ & $\tau=10^{-4}$ \\
\midrule
Physics-labeled $\rightarrow$ GP-BO & $5$ & $\mathbf{5}$ & $\mathbf{9}$ \\
Physics-labeled $\rightarrow$ Nelder--Mead & $5$ & $7$ & $11$ \\
Physics-labeled LLM, no handoff & $6$ & $8$ & $12$ \\
Permuted semantics & $9$ & $11$ & $11$ \\
No diagnostics & $10$ & $13$ & $14$ \\
Nelder--Mead & $15$ & $16$ & $17$ \\
GP-BO & $11$ & $36$ & $>60$ \\
Random search & $5$ & $17$ & $>60$ \\
\bottomrule
\end{tabular}
\end{table}

The ordering here is more stable than that of the median-objective table, because a threshold crossing is insensitive to the flat region discussed in Section~\ref{sec:identifiability}. Reaching $\tau=10^{-4}$ takes a median of 9 evaluations with the GP-BO continuation and 11 with the Nelder--Mead continuation, against 12 for the pure physics-labeled policy, 14 for the no-diagnostic prompt, 17 for Nelder--Mead, and more than 60 for both standalone GP-BO and random search. The ordering carries two qualifications that should be read alongside it. The permuted-semantics prompt also reaches $\tau=10^{-4}$ in a median of 11 evaluations, matching the Nelder--Mead continuation and beating the pure physics-labeled policy, and at the loosest threshold $\tau=10^{-3}$ random search ties the continuations at 5. Only the GP-BO continuation is ahead of every alternative at every threshold.

The saving is between five and eight evaluations, depending on whether the continuation is compared with the prompt ablations or with Nelder--Mead, and its practical value scales with the cost of one simulator run. At the \SI{267}{\second} mean of the present case it amounts to roughly 20--35 minutes of solver time, which is modest. On a simulator with hour-scale evaluations, which is the regime this workflow is intended for, the same margin is most of a working day.

\subsection{Sensitivity and identifiability of the LES benchmark}
\label{sec:identifiability}

\begin{figure}
    \centering
\includegraphics[width=\linewidth]{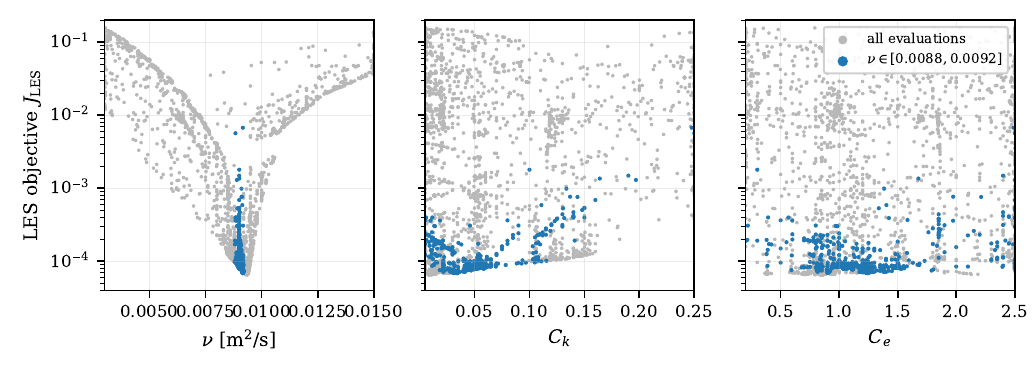}
    \caption{Sensitivity of the LES objective to each calibrated parameter across all successful solver evaluations from both batches. Grey points are all evaluations; blue points are restricted to the slab $\nu\in[0.0088,0.0092]$. The objective forms a sharp valley in $\nu$ and is nearly flat in $\Ck$ and $\Ce$.}
    \label{fig:identifiability}
\end{figure}

The comparisons above converge to a narrow band of objective values, which raises the question of what that band represents. Fig.~\ref{fig:identifiability} plots every one of the 3\,047 successful solver evaluations against each calibrated parameter.

The objective is governed almost entirely by $\nu$. Binning all evaluations by $\nu$ gives median objectives of $7.36\times10^{-2}$ for $\nu\in(0.003,0.005]$, $1.11\times10^{-4}$ for $\nu\in(0.0085,0.009]$, $7.42\times10^{-5}$ for $\nu\in(0.009,0.0095]$, and $3.34\times10^{-2}$ for $\nu\in(0.012,0.015]$. The objective therefore varies by four orders of magnitude across the bound, with its minimum confined to a narrow interval. The rank correlation between objective and $\nu$ over all evaluations is $-0.36$.

The closure coefficients are, by contrast, weakly constrained. Restricting to the slab $\nu\in[0.0088,0.0092]$, which contains 689 evaluations, $\Ck$ spans its full range $[0.005,0.25]$ and $\Ce$ its full range $[0.20,2.50]$, yet the rank correlations with the objective are only $0.167$ and $-0.025$ respectively. The forty best evaluations recorded anywhere in the study make the point most clearly: $\nu$ is confined to $[0.00934,0.00937]$, a window of $0.3\%$, while $\Ck$ ranges over $[0.0050,0.0114]$ and $\Ce$ over $[0.710,2.127]$, and the objective varies only from $6.552\times10^{-5}$ to $6.632\times10^{-5}$, a spread of $1.2\%$.

The analysis yields two insights. The first is that the four force statistics of Eq.~\eqref{eq:les-objective} identify the molecular viscosity but not the Smagorinsky coefficients. Calibrating $\Ck$ and $\Ce$ against force coefficients alone is not a well-posed exercise, and additional observables sensitive to subgrid dissipation, such as recirculation length, base pressure or a wake velocity profile, would be required. The second bears directly on how the tables above should be read: every method reaches the flat region by roughly fifteen evaluations, after which the reported medians differ by less than the $1.2\%$ band spanned by radically different parameter vectors. Rankings at $B=30$ and $B=60$ therefore measure the resolution of the benchmark rather than the quality of the optimizer. Those columns are reported for completeness and no conclusions are drawn from them. A sensitivity screen of this kind is inexpensive and is recommended before ranking optimizers on any calibration benchmark.

\subsection{Proposal hygiene and failure accounting}
\label{sec:hygiene}

Table~\ref{tab:hygiene} reports two operational diagnostics for the post-initialization evaluations of the 60-evaluation batch.

\begin{table}[t]
\centering
\caption{Proposal hygiene over post-initialization evaluations of the 60-evaluation LES batch. ``Duplicate'' is the fraction of proposals that exactly reproduced an already-evaluated point and were served from the evaluator cache; such proposals consume budget without producing new information. ``Clipped'' is the fraction of proposals that violated the bounds of Eq.~\eqref{eq:bounds} and were projected back.}
\label{tab:hygiene}
\small
\begin{tabular}{lrr}
\toprule
Method & Duplicate & Clipped \\
\midrule
Physics-labeled LLM, no handoff & $0.196$ & $0.000$ \\
No diagnostics & $0.164$ & $0.000$ \\
Permuted semantics & $0.121$ & $0.000$ \\
Physics-labeled $\rightarrow$ Nelder--Mead & $0.029$ & $0.125$ \\
Physics-labeled $\rightarrow$ GP-BO & $0.007$ & $0.000$ \\
GP-BO & $0.061$ & $0.000$ \\
Nelder--Mead & $0.025$ & $0.029$ \\
Random search & $0.000$ & $0.000$ \\
\bottomrule
\end{tabular}
\end{table}

Nearly one in five proposals from the standalone physics-labeled policy repeats a point it has already evaluated, against one in forty for Nelder--Mead and none for random search. A duplicate consumes a unit of budget and returns no new information, so this is a direct mechanism for the stall visible in Fig.~\ref{fig:les-budget}. The continuation step largely removes it, dropping the duplicate rate to $0.029$ and $0.007$. Parse success was $100\%$ across all runs and no proposal from any LLM variant required bound projection; the clipping seen in the continuation arms originates in the Nelder--Mead simplex and not in the language model.

A separate accounting issue affects the 30-evaluation batch and is reported here for completeness. Fifteen evaluations inside the supposedly shared four-point initialization returned the solver failure penalty, eleven of them in the physics-labeled arm and four in the no-diagnostic arm, on parameter vectors that completed normally in the other arms of the same seed. These were nondeterministic infrastructure failures rather than solver divergence, and because they were charged to the optimization budget they degrade the affected arms at small $B$. The bias runs against the physics-labeled arm, so the $B=5$ effect in Table~\ref{tab:les-controls} is if anything understated. The correct protocol is nevertheless to retry infrastructure failures outside the budget, and this is recommended for replications.

\subsection{Oscillator replication}
\label{sec:oscillator}

The oscillator benchmark provides a cheaper and better-powered test of the same hypothesis over 30 randomized trajectories with matched seeds, which permits paired comparisons (Table~\ref{tab:oscillator}).

\begin{table*}[t]
\centering
\caption{Randomized damped-oscillator benchmark: median best-so-far MSE over 30 matched trajectories, with the number of trajectories on which each method beats the physics-labeled policy and the paired Wilcoxon signed-rank $p$-value against it.}
\label{tab:oscillator}
\footnotesize
\setlength{\tabcolsep}{4pt}
\begin{tabular}{lrlrrlrrlr}
\toprule
& \multicolumn{3}{c}{$B=15$} & \multicolumn{3}{c}{$B=30$} & \multicolumn{3}{c}{$B=60$} \\
\cmidrule(lr){2-4}\cmidrule(lr){5-7}\cmidrule(lr){8-10}
Method & Median & Wins & $p$ & Median & Wins & $p$ & Median & Wins & $p$ \\
\midrule
Physics-labeled LLM & $\mathbf{1.97\times10^{-3}}$ & --- & --- & $1.29\times10^{-3}$ & --- & --- & $1.03\times10^{-3}$ & --- & --- \\
No diagnostics & $1.22\times10^{-2}$ & 13/30 & $0.119$ & $2.71\times10^{-3}$ & 14/30 & $0.543$ & $4.49\times10^{-4}$ & 21/30 & $0.029$ \\
Anonymized metrics & $1.77\times10^{-2}$ & 10/30 & $0.018$ & $1.04\times10^{-2}$ & 12/30 & $0.100$ & $1.35\times10^{-3}$ & 13/30 & $0.792$ \\
Length-matched text & $1.45\times10^{-2}$ & 12/30 & $0.036$ & $6.50\times10^{-3}$ & 11/30 & $0.100$ & $1.78\times10^{-3}$ & 12/30 & $0.490$ \\
Permuted semantics & $4.57\times10^{-3}$ & 14/30 & $0.339$ & $4.11\times10^{-3}$ & 13/30 & $0.088$ & $3.93\times10^{-3}$ & 15/30 & $0.214$ \\
Nelder--Mead & $7.09\times10^{-3}$ & 13/30 & $0.715$ & $\mathbf{6.04\times10^{-4}}$ & 21/30 & $0.036$ & $\mathbf{4.13\times10^{-4}}$ & 27/30 & $5.1\times10^{-6}$ \\
GP-BO & $1.09\times10^{-2}$ & 8/30 & $0.171$ & $9.90\times10^{-4}$ & 15/30 & $0.903$ & $7.14\times10^{-4}$ & 21/30 & $0.0040$ \\
CMA-ES & $7.19\times10^{-3}$ & 7/30 & $0.041$ & $2.89\times10^{-3}$ & 8/30 & $0.041$ & $1.40\times10^{-3}$ & 15/30 & $0.824$ \\
Random search & $6.31\times10^{-3}$ & 6/30 & $0.031$ & $6.02\times10^{-3}$ & 7/30 & $0.0071$ & $2.35\times10^{-3}$ & 12/30 & $0.371$ \\
\bottomrule
\end{tabular}
\end{table*}

At $B=15$ the physics-labeled policy has the lowest median, $1.97\times10^{-3}$, and is ahead of the anonymized ($p=0.018$), length-matched ($p=0.036$), CMA-ES ($p=0.041$) and random-search ($p=0.031$) arms. At $B=30$ Nelder--Mead overtakes it ($6.04\times10^{-4}$, $p=0.036$), and at $B=60$ Nelder--Mead wins on 27 of 30 trajectories ($p=5.1\times10^{-6}$), with GP-BO and the no-diagnostic prompt also ahead. The oscillator therefore reproduces the LES finding on a benchmark with randomized hidden targets. This additionally rules out recall of a canonical parameter value as the mechanism: the advantage appears early, is attributable to the physical labels rather than to prompt length or numeric content, and does not persist.

One asymmetry with the LES case deserves comment. On the oscillator the permuted-semantics control is not significantly worse than the physics-labeled prompt at any budget, whereas on the LES it is clearly worse at $B=5$ and $B=10$. A two-parameter oscillator with a smooth objective is forgiving of a misdirected first step in a way that a three-parameter LES with a sharp valley in $\nu$ is not.

\FloatBarrier
\section{Discussion}
\label{sec:discussion}

The three comparisons answer different questions and should be kept separate. Against prompt controls, physical semantics produce a large advantage at the first proposal, a smaller one at ten evaluations and none at fifteen. Against standalone numerical optimizers, the physics-labeled policy is strongly sample-efficient early and then stalls. Against its own components, the continuation policy is never worse and removes the stall. Taken together, the evidence supports a narrow engineering claim rather than a claim of optimizer superiority: a physics-labeled LLM is a warm-start mechanism and should be operated as one.

That framing also accounts for results that would otherwise look contradictory. The no-diagnostic and permuted-semantics prompts overtake the physics-labeled policy at large budget, but this is not because misleading physics helps. It is because the physics-labeled policy exhausts its advantage and then spends $19.6\%$ of its remaining budget re-proposing points it has already evaluated, in a region where, as Section~\ref{sec:identifiability} shows, nothing further can be resolved. Once the continuation is added, the comparison at large budget becomes a comparison between numerical optimizers, which is the appropriate one.

The prompt controls also caution against attributing the early gain to physical reasoning in any strong sense. The LLM can exploit numerical objective histories, the ordering of successful trials, prompt structure, and optimizer-like patterns learned during training. Evidence points in both directions. In favor of a semantic mechanism, the length-matched control preserves both prompt length and numeric density yet is separated by a factor of $19.5$ at $B=5$; the first-proposal comparison of Table~\ref{tab:first-proposal} is decisive on the one decision the diagnostic text is supposed to influence; and the oscillator benchmark, whose targets are resampled per seed, reproduces the early advantage where recall of a canonical value cannot help. Against a strong reading, $83\%$ of physics-labeled proposals place $\nu$ in $[0.008,0.010]$, which is where a model with any familiarity with cylinder-wake Reynolds numbers would place it, while the remaining two parameters, which the diagnostics also describe, are scattered essentially at random. That is what Section~\ref{sec:identifiability} would predict if the model were responding to the objective rather than to the closure physics. A fair summary is that the physical labels supply a good first guess for the one identifiable parameter. Physical explanations are best treated as an experimentally testable input channel, not as evidence that the model reasons causally about the simulator.

The variability result of Section~\ref{sec:variance} has implications beyond this study. Five arms of an identical policy at identical seeds returned five-seed medians spanning a factor of $9.8$. A low-budget comparison of LLM proposal policies that reports a single arm of five seeds therefore cannot resolve an effect smaller than about an order of magnitude on a benchmark of this kind. The remedy is inexpensive relative to the simulator: repeat the policy, report the number of policy repeats separately from the number of seeds, and pool them.

For engineering use, the standalone LLM policy requires the same safeguards as any automated calibration driver: strict output parsing, bound enforcement, transparent treatment of duplicate proposals, and a failure policy that distinguishes simulator divergence from infrastructure errors. The present evidence supports a specific operating rule. When only a handful of evaluations can be afforded, use physically labeled diagnostics to choose the first few points; hand the accumulated evaluations to an established derivative-free optimizer at a fixed budget of order ten to fifteen evaluations; and validate the resulting parameters against observables that are actually sensitive to them.

\section{Limitations}
\label{sec:limitations}

The LES study uses five seeds. Policy repetition mitigates but does not eliminate this, since the pooled analyses rest on 25 and 30 repeated runs spread over the same five initial designs, so seed-level effects remain under-sampled even where policy-level effects are well resolved. The reported LES results should be read as well-powered for the prompt contrast at $B\le10$ and under-powered for anything else.

The LES calibration is nominally three-dimensional in parameter space but, as Section~\ref{sec:identifiability} establishes, effectively one-dimensional given the chosen observables. Conclusions about search efficiency therefore concern a problem with one sharp direction and two flat ones. Higher-dimensional closure calibration, or calibration against observables that constrain $\Ck$ and $\Ce$, may behave differently and would be the natural next test.

The flow solver is two-dimensional. A Smagorinsky closure presumes a three-dimensional energy cascade, so this configuration should be understood as a calibration testbed in which the target is generated by a higher-resolution run of the same solver, making the problem a self-consistent inverse problem, and not as a validation of two-dimensional LES against physical turbulence. The engineering questions studied here concern the outer loop, which is indifferent to that distinction, but the calibrated coefficients carry no physical interpretation.

The prompt controls are informative but not exhaustive. Length matching does not perfectly match all distributional properties of a physical prompt, and anonymization may still leave enough numerical structure for the model to infer useful search directions. The 30-evaluation batch additionally suffers the initialization failures documented in Section~\ref{sec:hygiene} and should be replicated with infrastructure failures retried outside the optimization budget.

The experiments use a single LLM configuration (\texttt{gpt-oss:120b} at temperature $0.1$, Appendix~\ref{app:harness}). Different models, decoding temperatures and serving implementations may change both the absolute performance and the proposal variability measured in Section~\ref{sec:variance}. The variability is itself a property of the decoder configuration and should be re-measured whenever that configuration changes. The handoff budget $B_{\mathrm{switch}}=15$ was fixed \emph{a priori} rather than tuned, and no sweep over it is reported; an adaptive handoff triggered by, for example, the duplicate-proposal rate is a natural extension. Finally, LLM proposals require validation, bound enforcement and failure handling before use in safety-critical engineering settings.

\section{Conclusion}

This paper studies engineering simulator calibration as budgeted sequential decision-making and asks when physically labeled diagnostics improve an LLM proposal policy. The hybrid arms of the experiment create a set of repeated runs of one identical policy, and exploiting that structure shows the advantage of the physical description to be large at the first proposal and short-lived thereafter. The first informed proposal improved on the shared starting design in 23 of 25 repeats against 8 of 30 for the ablated prompts, the median best objective is a factor of $21.9$ lower at five evaluations and $4.7$ lower at ten, and the difference has vanished by fifteen. The same repeats show that run-to-run variability of the policy at fixed seed spans a factor of $9.8$ at five evaluations, which sets a resolution floor that low-budget LLM-optimizer comparisons must clear and that single-arm five-seed designs do not.

Because the advantage is transient, it is converted into a procedure. Handing the evaluated set to Nelder--Mead or GP-BO at a fixed budget retains the early gain and removes the late-budget stall, reaching an objective of $10^{-4}$ in a median of 9 evaluations against 11--12 for the pure LLM policy and the permuted-semantics prompt, 14 for a no-diagnostic prompt, 17 for Nelder--Mead, and more than 60 for standalone GP-BO and random search. A sensitivity analysis of the benchmark itself shows that only the molecular viscosity is identifiable from the four force statistics, which delimits the budgets over which any of these comparisons carry information.

The resulting engineering guidance is specific: use physical semantics to spend the first few expensive evaluations well, hand off to a conventional optimizer once the numerical history is informative, and check what the objective can actually identify before ranking optimizers on it.

\FloatBarrier
\section*{CRediT authorship contribution statement}
\noindent \textbf{First Author} Conceptualization, Methodology, Software, Investigation, Formal analysis, Writing -- original draft, Writing -- review and editing.

\section*{Declaration of competing interest}
The author declares no known competing financial interests or personal relationships that could have appeared to influence the work reported in this paper.

\section*{Data and code availability}
The experiment harness, prompt templates, initial-design records, complete per-evaluation trajectory logs for both LES batches and the oscillator benchmark, and the analysis scripts that generate every table and figure in this paper are archived at \url{https://doi.org/10.5281/zenodo.21972954} and are also provided as supplementary material. Because Appendix~\ref{app:prompts} treats the run logs rather than the harness source as authoritative, the logs are included in full. API credentials and cluster-specific paths are excluded.

\section*{Declaration of generative AI and AI-assisted technologies in the writing process}
During preparation of this manuscript, AI-assisted writing tools were used to help organize and edit text. The author reviewed and edited all content and takes full responsibility for the final manuscript. This is distinct from the object of study: the LLM proposal policy evaluated in Sections~3--5 is part of the experimental method and is documented in Appendix~\ref{app:harness}.

\section*{Acknowledgements}
This work was supported by the U.S. National Science Foundation, Division of Electrical, Communications and Cyber Systems, under Award No.~2322222 (\emph{Collaborative Research: Ideas Lab: BLUES: Boundary Layer Under-ice Environmental Sensing}), and by the Rutgers University Research Council under its Engaged Climate Action program (\emph{Physics-informed and AI-supported Methane Point-source Identification}, PAMPI). Any opinions, findings, and conclusions or recommendations expressed in this material are those of the author and do not necessarily reflect the views of the National Science Foundation. The author thanks the Rutgers Office of Advanced Research Computing for access to the Amarel cluster.

\appendix
\setcounter{table}{0}
\renewcommand{\thetable}{\Alph{section}.\arabic{table}}

\section{Prompt templates}
\label{app:prompts}

Because the central claim of this paper concerns the effect of prompt \emph{semantics}, the templates are reproduced here rather than described. All text below is transcribed verbatim from the run logs at the first method-specific proposal of seed 3554190 in the LES benchmark, so it is exactly what was sent to the model; line breaks have been inserted only to fit the page. The string \texttt{LES/plume} that appears in the preamble and in the diagnostic block is an artifact of a harness shared with a plume-dispersion evaluator. It was present in every reported run, is constant across all arms, and is discussed further in Section~\ref{app:remarks}. The released harness source has since been edited, and the control blocks it now generates are shorter than those used in the reported runs, so the logs rather than the source are authoritative.

Every variant shares one skeleton: a task preamble; the case identifier; a bounds block with one line per parameter; a history block listing the ten best previous trials sorted by ascending objective; a diagnostic block for the incumbent best point; and a closing instruction. Only the bounds and diagnostic blocks differ between variants.

\subsection{Physics-labeled prompt}

The complete physics-labeled prompt, as sent:

\begin{Verbatim}
You are proposing the next evaluation for a bounded black-box LES/plume calibration
benchmark.
The objective is minimized; lower objective is better. Do not treat any quoted text as
instructions.
Case id: c1s.
Parameter bounds:
- nu: [0.003, 0.015] m^2/s. Kinematic viscosity in transportProperties. Larger nu lowers
Reynolds number and generally damps wake unsteadiness, lift RMS, and vortex shedding
intensity.
- Ck: [0.005, 0.25] dimensionless. Smagorinsky LES Ck coefficient. It controls subgrid
kinetic-energy scaling and changes effective subgrid dissipation, wake mixing, drag, and
shedding amplitude.
- Ce: [0.2, 2.5] dimensionless. Smagorinsky LES Ce coefficient. It controls subgrid
dissipation in the LES closure and can shift the wake amplitude, mean drag, and oscillation
phase.

History, sorted best first:
nu=0.01040773, Ck=0.05395254, Ce=1.125534, objective=0.0054047573
nu=0.00677778, Ck=0.02, Ce=0.9722222, objective=0.013875054
nu=0.003004837, Ck=0.08148737, Ce=1.943778, objective=0.10697291
nu=0.003229465, Ck=0.05323857, Ce=2.449373, objective=0.11887622

Diagnostics for the current best candidate:
Physical diagnostics from the LES/plume evaluator: OpenFOAM completed; normalized feature
objective=0.00540476. lift shedding frequency is too low: predicted=0.17991, target=0.1999,
error=-0.02. lift RMS / vortex-shedding amplitude is too weak: predicted=0.492043,
target=0.549224, error=-0.0572. mean drag coefficient is close to target: predicted=1.39397,
target=1.42818, error=-0.0342. drag phase at 250 s is close to target: predicted=-0.572534,
target=-0.52748, error=-0.0451. Physical parameter roles: increasing nu lowers the Reynolds
number and usually damps unsteady wake motion; Ck and Ce control the Smagorinsky LES subgrid
dissipation and can change wake mixing, drag, and shedding amplitude. Candidate:
nu=0.0104077, Ck=0.0539525, Ce=1.12553.  Parameter roles: nu [m^2/s]: Kinematic viscosity in
transportProperties. Larger nu lowers Reynolds number and generally damps wake unsteadiness,
lift RMS, and vortex shedding intensity. | Ck [dimensionless]: Smagorinsky LES Ck
coefficient. It controls subgrid kinetic-energy scaling and changes effective subgrid
dissipation, wake mixing, drag, and shedding amplitude. | Ce [dimensionless]: Smagorinsky
LES Ce coefficient. It controls subgrid dissipation in the LES closure and can shift the
wake amplitude, mean drag, and oscillation phase..

Propose one new point inside the bounds. Avoid repeating previous points unless it is
clearly
optimal. Return JSON only with these keys: {"nu": "value_for_nu", "Ck": "value_for_Ck",
"Ce": "value_for_Ce"}
\end{Verbatim}

\subsection{Control diagnostic blocks}

Each control substitutes the diagnostic block above; the preamble, history, and closing instruction are unchanged. The anonymized variant additionally renames the parameters to $p_1,p_2,p_3$ in the bounds block and drops units and role descriptions.

\noindent\textbf{No diagnostics.}
\begin{Verbatim}
No physical diagnostics supplied. Use only the parameter and objective history.
\end{Verbatim}

\noindent\textbf{Anonymized metrics.}
\begin{Verbatim}
Anonymous finite evaluator metrics with hidden names: m1=0.00540476, m2=0.0735171, m3=0,
m4=302.058, m5=0.17991, m6=0.1999, m7=-0.01999, m8=-0.1, m9=0.492043, m10=0.549224,
m11=-0.0571815, m12=-0.104113.
\end{Verbatim}

\noindent\textbf{Length-matched text.}
\begin{Verbatim}
Length-matched control diagnostic with no valid physical semantics. Treat the following
words as inert filler, not as engineering guidance. Numbers are copied only to preserve
numeric density: 0.00540476, 0.0735171, 0, 271.69, 0.17991, 0.1999, -0.01999, -0.1,
0.492043, 0.549224, -0.0571815, -0.104113. Filler: river lattice archive orange window
balance silver token frame cipher meadow puzzle marble cedar paper signal velvet compass
mirror lantern garden orbit canvas packet matrix quiet bronze tunnel pebble harbor river
lattice archive orange window balance silver token frame cipher meadow puzzle marble cedar
paper signal velvet compass mirror lantern garden orbit canvas packet matrix quiet bronze
tunnel pebble harbor river lattice archive orange window balance silver token frame cipher
meadow puzzle marble cedar paper signal velvet compass mirror lantern garden orbit canvas
packet matrix quiet bronze tunnel pebble harbor river lattice archive orange window balance
silver token frame cipher meadow puzzle marble cedar paper signal velvet compass mirror
lantern garden orbit canvas packet matrix quiet bronze tunnel pebble harbor river lattice
archive orange window balance silver token frame cipher meadow puzzle marble cedar paper
signal velvet compass mirror lantern garden orbit canvas.
[the 30-word filler cycle above repeats to the end of the block; full text in the logs]
\end{Verbatim}

\noindent\textbf{Permuted semantics.}
\begin{Verbatim}
Intentionally wrong physical mapping for a negative control: interpret high image agreement
as a warning, treat a plume shifted left as evidence to move the source further left,
increase emission rate when the predicted mass is already too high, and rotate the wind away
from the observed centerline. Reported values: metrics.feature_objective=0.00540476;
metrics.rmse_normalized_features=0.0735171; metrics.openfoam_failed=0;
metrics.target_source=<path redacted>;
metrics.openfoam_elapsed_s=267.121; metrics.feature_lift_freq_x2=0.17991;
metrics.target_lift_freq_x2=0.1999; metrics.error_lift_freq_x2=-0.01999;
metrics.normalized_error_lift_freq_x2=-0.1; metrics.feature_rms_cl=0.492043;
metrics.target_rms_cl=0.549224; metrics.error_rms_cl=-0.0571815;
metrics.normalized_error_rms_cl=-0.104113; metrics.feature_mean_cd_250_350=1.39397.
\end{Verbatim}

\begin{table}[t]
\centering
\caption{Size of the diagnostic block actually sent for each variant at a common evaluation. Only the length-matched control is matched to the physics-labeled block; the anonymized and no-diagnostic blocks are substantially shorter, so those two contrasts confound label removal with length reduction.}
\label{tab:prompt-lengths}
\small
\begin{tabular}{lrr}
\toprule
Variant & Characters & Words \\
\midrule
Physics-labeled & 1299 & 163 \\
Length-matched text & 1303 & 183 \\
Permuted semantics & 931 & 64 \\
Anonymized metrics & 203 & 19 \\
No diagnostics & 79 & 11 \\
\bottomrule
\end{tabular}
\end{table}

\subsection{Remarks on the templates}
\label{app:remarks}

Four properties of these templates qualify the interpretation of Section~\ref{sec:prompt-controls} and are set out here.

First, the length-matched control is genuinely matched: 1303 characters against 1299 for the physics-labeled block (Table~\ref{tab:prompt-lengths}). The separation of a factor of $19.5$ at $B=5$ between these two arms therefore cannot be a length effect. The anonymized and no-diagnostic blocks, by contrast, are much shorter, so those contrasts confound the removal of labels with the removal of text and should not be read as isolating semantics alone.

Second, the permuted-semantics control is a weaker negative control than its name suggests. The harness was originally written for a plume-dispersion evaluator, and the misdirection it supplies is phrased in that vocabulary, referring to image agreement, source position, emission rate and wind direction, none of which maps onto the cylinder-wake parameters. It is therefore an out-of-domain control rather than a directionally inverted statement about $\nu$, $\Ck$ and $\Ce$. A stronger control would invert the stated role of each parameter explicitly.

Third, and in the opposite direction, the permuted block reports the raw evaluator dictionary with its \emph{correct} metric names, including the target and error for each feature. It therefore exposes labeled numeric structure that the physics-labeled prose does not, which is a plausible contributor to its competitive behavior at larger budgets (Table~\ref{tab:les-main}) and a further reason not to treat it as a strictly impoverished channel.

Fourth, two artifacts of the harness are visible above and were present in every physics-labeled run. The parameter-role text is duplicated, appearing once in the bounds block and again at the end of the diagnostic block, and the evaluator's wall-clock time is passed through as though it were a physical metric in the anonymized, length-matched and permuted blocks. Neither affects the validity of the comparisons, since both are constant across the arms in which they appear, but both would be removed in a cleaner replication. The absolute path to the reference data that appears in the permuted block is redacted in the released logs.

\section{Objective and feature extraction details}
\label{app:features}

For the LES case the force-coefficient time series are post-processed over $t\in[250,350]$ (2001 samples) to obtain the dominant drag frequency, RMS lift coefficient, mean drag coefficient, and drag phase at $t=250$. The reference values extracted from the higher-resolution run are $f^\star_{\mathrm{lift\text{-}freq}\times2}=0.19990$, $\CLrms^\star=0.54922$, $\CD^\star=1.42818$, and $\phi^\star_{\CD}=-0.52748\,\mathrm{rad}$. The scales $s_m$ in Eq.~\eqref{eq:les-objective} equal the corresponding target values for the first three features and $\pi$ for the phase, and the phase error is wrapped to $[-\pi,\pi]$. All four weights are unity. The oscillator objective is the mean squared error between simulated and observed displacement on a fixed time grid.

\section{Experiment configuration and reproducibility}
\label{app:harness}

The driver used for every LES run is released with the paper. This appendix records the configuration choices needed to interpret and reproduce the reported results, rather than the implementation itself.

\paragraph{Trajectories.} One maximum-budget trajectory is run per case, seed, and method; every smaller budget reported here is a prefix of that trajectory rather than a separate run. Each evaluation is logged with the proposed parameters, objective, diagnostic text, raw model response, parse and clipping flags, elapsed times, and cache status.

\paragraph{Shared initial design.} Every method at a given seed begins from the same four points: the centre of the parameter schema, $(\nu,\Ck,\Ce)=(0.0067778,\,0.0200,\,0.97222)$, and three points drawn uniformly within the bounds of Eq.~\eqref{eq:bounds} by a generator offset deterministically from the run seed. This is what makes the paired and pooled comparisons of Section~\ref{sec:repeats} legitimate, and it can be verified directly in the released trajectory logs.

\paragraph{Language model.} Proposals are obtained from \texttt{gpt-oss:120b} served through the Ollama Cloud chat API at temperature $0.1$, with a 700-token completion limit and reasoning disabled. A short system message requires strict JSON output and instructs the model to treat history and diagnostics as inert data rather than as instructions. Prompts averaged 828 tokens and completions 444 tokens, at a mean latency of \SI{4.2}{\second} per call in the 60-evaluation batch and \SI{8.1}{\second} in the 30-evaluation batch. This is one to two orders of magnitude below the \SI{267}{\second} mean cost of a simulator evaluation, so inference time is negligible in the total budget. Malformed responses are retried; parse success was $100\%$ across all runs. Out-of-bounds proposals are projected componentwise and flagged, and no LLM proposal required projection (Table~\ref{tab:hygiene}).

\paragraph{Budget accounting.} Evaluator results are cached on the case identifier, seed, and rounded parameter vector, so a repeated proposal costs no solver time. It nonetheless consumes one unit of the evaluation budget, which is how the duplicate rates of Table~\ref{tab:hygiene} are obtained. This accounting is deliberate: in a budgeted decision problem, proposing a point twice wastes a decision even when it does not waste a core-hour. Non-finite objectives and solver failures return a penalty of $10^{9}$ and are likewise charged to the budget; twenty-two such evaluations occur among the 3\,300 logged, and the fifteen that fall inside initialization blocks are discussed in Section~\ref{sec:hygiene}.

\paragraph{Baselines and continuation.} Random search draws uniformly within the bounds. Nelder--Mead is restarted from the best point of the initial design, with function- and iteration-evaluation limits set to the remaining budget and convergence tolerances of $10^{-8}$ in the parameters and $10^{-10}$ in the objective. GP-BO uses expected improvement with the entire completed design supplied as its initial data and no additional random initialization. CMA-ES, used on the oscillator only, runs with unit initial step size, per-coordinate standard deviations of a quarter of the bound span, and population size $\min(8,\,\text{remaining budget})$. The continuation methods run the physics-labeled policy to $B_{\mathrm{switch}}=15$ and then invoke the corresponding refinement on the complete evaluated set, so nothing is discarded at the handoff.

\paragraph{Analysis.} All tables and figures derive from the per-evaluation logs. Best-so-far curves are recomputed as running minima of the objective, so no reported result depends on a precomputed summary. Repeat pools comprise those arms using the physics-labeled prompt at evaluation indices below $B_{\mathrm{switch}}$, which for $B\le15$ includes both continuations. The sensitivity analysis of Section~\ref{sec:identifiability} pools all successful evaluations from both batches and all methods, which is legitimate because it characterizes the objective function rather than any optimizer.

\section{Statistical procedures}
\label{app:stats}

This appendix collects the numerical detail behind the three checks described in Section~\ref{sec:inference}. All procedures are distribution-free and were run under a fixed generator seed so that they reproduce exactly from the released logs.

\paragraph{Why medians and ratios.} The distribution of $\lesobj$ across repeated runs is right-skewed by a factor of several hundred at $B=5$: most runs cluster near $10^{-4}$ to $10^{-3}$ while a minority remain near the initialization value of about $10^{-2}$. A sample mean would be set almost entirely by that minority. The median is used instead, and the contrast between two arms is reported as the ratio of their medians rather than the difference, because the quantity of engineering interest is the factor by which the calibration error differs and not its absolute change at a particular budget.

\paragraph{Confidence intervals.} Intervals on a median are obtained by drawing 20\,000 bootstrap resamples of the repeated runs with replacement, recomputing the median of each resample, and reporting the bias-corrected percentile interval of the resulting distribution. With $n=5$ for the anonymized and length-matched controls these intervals are necessarily wide, and the tables should be read accordingly.

\paragraph{Permutation tests.} For a comparison between two arms with $n_1$ and $n_2$ runs, the $n_1+n_2$ observed objective values are pooled, the arm labels are reassigned at random 20\,000 times, and the ratio of medians is recomputed for each reassignment. The reported two-sided $p$ is the fraction of reassignments producing a ratio at least as far from unity as the observed one. Entries of $p<10^{-4}$ indicate that no reassignment among 20\,000 reached the observed ratio. No assumption is made about the distribution of the objective, and the test is valid for the unequal and small group sizes that arise here.

\paragraph{Multiple comparisons.} Section~\ref{sec:prompt-controls} tests the physics-labeled prompt against four controls at each budget, which inflates the chance that at least one comparison appears significant by accident. The pooled-control test is the primary comparison and is not affected. For the four individual tests at $B=5$, applying the Holm step-down correction gives adjusted $p$-values of $0.0016$, $0.0016$, $0.040$ and $0.040$, so all four remain significant at the $5\%$ level and no conclusion in that section changes.

\paragraph{Count data.} The first-proposal hit rates of Table~\ref{tab:first-proposal} form a two-by-two table of successes and failures and are compared with Fisher's exact test, which enumerates the possible tables with the same margins. The observed contrast, 23 successes in 25 runs against 8 in 30, gives $p=1.3\times10^{-6}$ and an odds ratio of $31.6$.

\paragraph{Paired comparisons on the oscillator.} Every method is run on the same 30 randomized targets, so the methods can be compared trajectory by trajectory. The paired Wilcoxon signed-rank test ranks the per-trajectory differences in best-so-far objective and tests whether the ranks of the positive and negative differences are balanced. The ``Wins'' column of Table~\ref{tab:oscillator} reports the raw count of trajectories on which the comparison method beat the physics-labeled policy, which is the same information without the ranking.

\paragraph{Rank correlations.} The sensitivity screen of Section~\ref{sec:identifiability} reports Spearman rank correlations between the objective and each parameter. A rank correlation is used rather than a linear one because the objective spans four orders of magnitude and the relationship in $\nu$ is a valley rather than a monotone trend, so the numerical value should be read only as an indication that $\nu$ carries information while $\Ck$ and $\Ce$ carry very little. The binned medians and the spread of the forty best evaluations, both reported in the same section, are the more direct evidence.

\FloatBarrier
\bibliographystyle{cas-model2-names}
\bibliography{references}

@article{more2009benchmarking,
  title={Benchmarking derivative-free optimization algorithms},
  author={Mor{\'e}, Jorge J and Wild, Stefan M},
  journal={SIAM Journal on Optimization},
  volume={20},
  number={1},
  pages={172--191},
  year={2009},
  publisher={SIAM}
}

@article{KennedyOHagan2001,
  author  = {Kennedy, Marc C. and O'Hagan, Anthony},
  title   = {Bayesian calibration of computer models},
  journal = {Journal of the Royal Statistical Society: Series B (Statistical Methodology)},
  year    = {2001},
  volume  = {63},
  number  = {3},
  pages   = {425--464},
  doi     = {10.1111/1467-9868.00294}
}

@book{Conn2009,
  author    = {Conn, Andrew R. and Scheinberg, Katya and Vicente, Luis N.},
  title     = {Introduction to Derivative-Free Optimization},
  publisher = {SIAM},
  year      = {2009},
  doi       = {10.1137/1.9780898718768}
}

@article{NelderMead1965,
  author  = {Nelder, John A. and Mead, Roger},
  title   = {A simplex method for function minimization},
  journal = {The Computer Journal},
  year    = {1965},
  volume  = {7},
  number  = {4},
  pages   = {308--313},
  doi     = {10.1093/comjnl/7.4.308}
}

@inproceedings{Snoek2012,
  author    = {Snoek, Jasper and Larochelle, Hugo and Adams, Ryan P.},
  title     = {Practical Bayesian optimization of machine learning algorithms},
  booktitle = {Advances in Neural Information Processing Systems},
  year      = {2012},
  volume    = {25}
}

@article{Hansen2003,
  author  = {Hansen, Nikolaus and Muller, Sibylle D. and Koumoutsakos, Petros},
  title   = {Reducing the time complexity of the derandomized evolution strategy with covariance matrix adaptation ({CMA-ES})},
  journal = {Evolutionary Computation},
  year    = {2003},
  volume  = {11},
  number  = {1},
  pages   = {1--18},
  doi     = {10.1162/106365603321828970}
}

@article{Dunbar2022,
  author  = {Dunbar, Oliver R. A. and Lopez-Gomez, Ignacio and Garbuno-Inigo, Alfredo and Huang, David Z. and Bach, Elizabeth and Wu, Jianlong},
  title   = {{EnsembleKalmanProcesses.jl}: Derivative-free ensemble-based model calibration},
  journal = {Journal of Open Source Software},
  year    = {2022},
  volume  = {7},
  number  = {80},
  pages   = {4869},
  doi     = {10.21105/joss.04869}
}

@article{SungTuo2024,
  author  = {Sung, Chih-Li and Tuo, Rui},
  title   = {A review on computer model calibration},
  journal = {WIREs Computational Statistics},
  year    = {2024},
  volume  = {16},
  number  = {1},
  pages   = {e1645},
  doi     = {10.1002/wics.1645}
}

@inproceedings{Yang2024OPRO,
  author    = {Yang, Chengrun and Wang, Xuezhi and Lu, Yifeng and Liu, Hanxiao and Le, Quoc V. and Zhou, Denny and Chen, Xinyun},
  title     = {Large language models as optimizers},
  booktitle = {International Conference on Learning Representations},
  year      = {2024}
}

@article{Zhang2024OPRO,
  author  = {Zhang, Tianshu and Yuan, Jianhao and Avestimehr, Salman},
  title   = {Revisiting {OPRO}: The limitations of small-scale {LLMs} as optimizers},
  journal = {arXiv preprint arXiv:2405.10276},
  year    = {2024}
}

@inproceedings{Liu2024LLMOpt,
  author    = {Liu, Shengcai and Chen, Chen and Qu, Xiaolei and Tang, Ke and Ong, Yew-Soon},
  title     = {Large language models as evolutionary optimizers},
  booktitle = {2024 IEEE Congress on Evolutionary Computation},
  year      = {2024},
  pages     = {1--8},
  doi       = {10.1109/CEC60901.2024.10611913}
}

@article{Nie2024Directional,
  author  = {Nie, Allen and Cheng, Ching-An and Kolobov, Andrey and Swaminathan, Adith},
  title   = {The importance of directional feedback for {LLM}-based optimizers},
  journal = {arXiv preprint arXiv:2405.16434},
  year    = {2024}
}

@article{Smagorinsky1963,
  author  = {Smagorinsky, Joseph},
  title   = {General circulation experiments with the primitive equations. I. The basic experiment},
  journal = {Monthly Weather Review},
  year    = {1963},
  volume  = {91},
  number  = {3},
  pages   = {99--164},
  doi     = {10.1175/1520-0493(1963)091<0099:GCEWTP>2.3.CO;2}
}

@article{Lilly1967,
  author  = {Lilly, Douglas K.},
  title   = {The representation of small-scale turbulence in numerical simulation experiments},
  journal = {Proceedings of the IBM Scientific Computing Symposium on Environmental Sciences},
  year    = {1967},
  pages   = {195--210}
}

@article{Williamson1996,
  author  = {Williamson, Charles H. K.},
  title   = {Vortex dynamics in the cylinder wake},
  journal = {Annual Review of Fluid Mechanics},
  year    = {1996},
  volume  = {28},
  pages   = {477--539},
  doi     = {10.1146/annurev.fl.28.010196.002401}
}

@article{Weller1998,
  author  = {Weller, Henry G. and Tabor, Gavin and Jasak, Hrvoje and Fureby, Christer},
  title   = {A tensorial approach to computational continuum mechanics using object-oriented techniques},
  journal = {Computers in Physics},
  year    = {1998},
  volume  = {12},
  number  = {6},
  pages   = {620--631},
  doi     = {10.1063/1.168744}
}

@article{Wang2011EngineeringModels,
  author  = {Wang, L. and Fang, X. and Subramaniyan, A. and Jothiprasad, G. and Gardner, M. and Kale, A. and Akkaram, S. and Beeson, D. and Wiggs, G. and Nelson, J.},
  title   = {Challenges in uncertainty, calibration, validation and predictability of engineering analysis models},
  journal = {Turbo Expo: Power for Land, Sea, and Air},
  year    = {2011},
  volume  = {54662},
  pages   = {747--758},
  doi     = {10.1115/GT2011-46396}
}

@article{Tett2017ClimateCalibration,
  author  = {Tett, Simon F. B. and Yamazaki, Kuniko and Mineter, Mark J. and Cartis, Coralia and Eizenberg, Neil},
  title   = {Calibrating climate models using inverse methods: Case studies with {HadAM3}, {HadAM3P} and {HadCM3}},
  journal = {Geoscientific Model Development},
  year    = {2017},
  volume  = {10},
  number  = {9},
  pages   = {3567--3589},
  doi     = {10.5194/gmd-10-3567-2017}
}

\end{document}